\documentclass[final]{cvpr}
\usepackage{caption} 
\usepackage[dvipsnames]{xcolor}
\usepackage{xcolor}
\usepackage{multirow}
\usepackage{times}
\usepackage{epsfig}
\usepackage{graphicx}
\usepackage{amsmath}
\usepackage{amssymb}
\usepackage{diagbox}

\usepackage[pagebackref=true,breaklinks=true,colorlinks,bookmarks=false]{hyperref}

\def\cvprPaperID{****} 
\def\confYear{CVPR 2021}

\begin{document}

\title{SeRP: Self-Supervised Representation Learning Using Perturbed Point Clouds}

\author{Siddhant Garg\\
University of Massachusetts Amherst \\
{\tt\small siddhantgarg@umass.edu}
\and
Mudit Chaudhary\\
University of Massachusetts Amherst\\
{\tt\small mchaudhary@umass.edu}
}

\maketitle

\begin{abstract}
   We present SeRP, a framework for Self-Supervised Learning of 3D point clouds. SeRP consists of encoder-decoder architecture that takes perturbed or corrupted point clouds as its inputs and aims to reconstruct the original point cloud without corruption. The encoder learns the high-level latent representations of the points clouds in a low-dimensional subspace and recover the original structure. In this work, we have used Transformers and PointNet based Autoencoders. The proposed framework also addresses some of the limitations of Transformers based Masked Autoencoders \cite{point-bert} which are prone to leakage of location information and uneven information density \cite{point-mae}. We pretrained our models on the complete ShapeNet dataset \cite{shapenet} and evaluated them on ModelNet40 \cite{modelnet40} as a downstream classification task. We have shown that  the pretrained models achieved $0.5-1\%$ higher classification accuracies than the networks trained from scratch. Furthermore, we also proposed VASP: \textbf{V}ector-Quantized \textbf{A}utoencoder for \textbf{S}elf-supervised Representation Learning for \textbf{P}oint Clouds  that employs Vector-Quantization \cite{vq-vae} for discrete representation learning for Transformer based autoencoders. 
\end{abstract}

\section{Introduction}
\label{sec:intro}

Self-Supervised learning methods have come a long way today. They have been extremely successful in the domains of Natural Language Processing \cite{devlin2018bert, electra} and 2D Computer Vision \cite{moco, byol} where large unlabeled datasets are leveraged to introduce necessary inductive biases in the models that can be transferred on the downstream tasks with limited annotated datasets. Recently, many self-supervised learning methods are also being proposed for 3D deep learning \cite{pointcontrast, Spatio-temporal-1, space-time-2, mmodel-1}, specifically for representation learning of the raw point clouds data. The annotation of point cloud is a difficult process and there are lesser annotated datasets in 3D deep learning relative to NLP and 2D computer vision. With the proliferation of modern 3D scanners, that can generate large unlabeled point clouds, 3D self-supervised learning methods is a promising direction to improve results on small annotated 3D datasets. 

\begin{figure}[t]
\begin{center}
   \includegraphics[width=\linewidth]{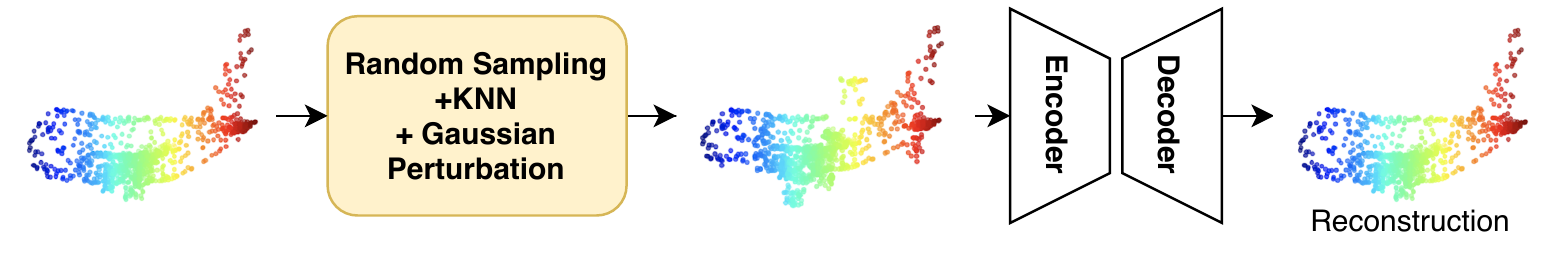}
\end{center}
   \caption{Overview of SeRP. First, an input point cloud is perturbed or corrupted and then it is processed by an autoencoder to learn latent representations using reconstruction process.}
\label{fig:overview}
\end{figure}

In this work we are proposing Auto-Encoders for learning high level latent representations of points clouds. As shown in Figure \ref{fig:overview}, the encoder takes a perturbed or noisy point cloud and the decoder aims to reconstruct the original point cloud from the latent representation. We used PointNet \cite{pointnet} and Point Cloud Transformers to design our Auto-Encoder framework. Our approach also overcomes the issue of information leakage associated with masked auto-encoding \cite{point-mae, point-bert}, where the position embeddings of the masks are also provided to the model \cite{point-mae, point-bert}.  

Our contributions can be summarized as follows:
\begin{itemize}
    \vspace{-1mm}
    \item Proposed a perturbation technique to create noisy point clouds by selecting small patches using random samplint and adding a Gaussian noise to the patch of point clouds.
    \vspace{-2mm}
    \item Proposed PointNet and Point Cloud Transformer based Auto-Encoders to learn latent representations of the point clouds. 
    \vspace{-2mm}
    \item Evaluated the pretrained encoder representations on downsteam classification tasks where we showed improvement of $0.5-1\%$ on ModelNet40 dataset and ShapeNet55 classification datasets. 
    \vspace{-2mm}
    \item Also proposed Vector-Quantization for Transformer based auto-encoding. 
\end{itemize}

\begin{figure*}
\begin{center}
\includegraphics[width=\linewidth]{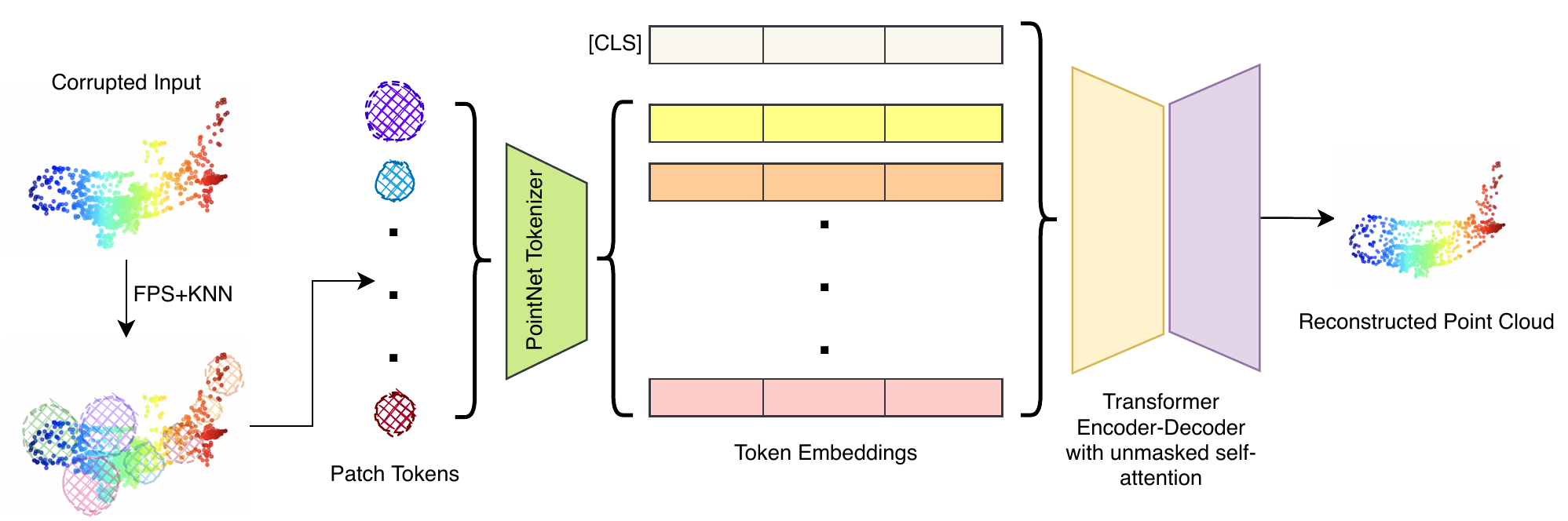}
\end{center}
  \caption{Overview of SeRP-Transformer model. The corrupted point cloud is divided into patches using FPS and KNN to create patch tokens. The token embeddings are learned using a lightweight PointNet. The embeddings are passed through a Transformer Encoder-Decoder with unmasked self-attention. The trained encoder is then used for downstream tasks.}
\label{fig:serp-transformer}
\end{figure*}

\section{Related Works}
\label{sec:related}
Self-supervised learning methods leverage the unlabeled data to learn robust representations. This can be done by developing pretext tasks like learning reconstruction spaces \cite{jigsaw}, orientation estimation \cite{orientation}, and occlusion completion \cite{occo} for point clouds. Some other techniques fall under the umbrella of contrastive \cite{pointcontrast, depthcontrast} and non-contrastive \cite{space-time-2, Spatio-temporal-1} learning methods that learn by making the representations of augmented views of same object, closer and augmented views of different objects farther from each other. 

Recently, a lot of work has been done towards 3D self-supervised representation learning from the raw point cloud data. Point Constrast \cite{pointcontrast} is a contrastive learning method that uses registration objective as a pretext task, DGCLR \cite{fu2022distillation} used knowledge distillation with constrastive learning, and recently, Point-BERT \cite{point-bert}, which is transformer \cite{devlin2018bert} based network used BERT-style pre-training by masking patches of point clouds and predict the masked parts with dVAE \cite{dvae}. Many auto-encoding methods like, Point-MAE \cite{point-mae}, and Implicit Autoencoders \cite{implicit-autoencoder} are also proposed that aim to reconstruct point clouds and surfaces respectively. Some of the recent works also try to leverage the spatio-temporal dimension of the input point clouds \cite{Spatio-temporal-1, space-time-2, space-time-3}. There also has been a lot of work to learn from multiple modalities \cite{mmodel-1, mmodel-2, mmodel-3}, like meshes and multi-view images, where the aim is to leverage information from all the modalities to encode the point clouds. Our work is based on auto-encoding \cite{point-mae} and reconstruction of point clouds \cite{point-bert} but existing works like Point-MAE \cite{point-mae}, and Point-BERT \cite{point-bert} used masking as a pretext task for transformer based encoders that are trained to reconstruct the masked point cloud patch. In contrast to the masking approach, we are corrupting the point cloud randomly and train our model to reconstruct the original point cloud.

\section{Methods}

\subsection{Point Cloud Perturbation}
To introduce robustness in the latent representations of the point clouds, we first corrupt the input point clouds before the decoder is trained to reconstruct the original point cloud. An input point cloud consists of a set of 1024 points representing a shape. To prevent the loss of the structure as well as introduce sufficient corruption, we selected 20 centers using random sampling and apply nearest neighbors to select 20 nearest points for each of the centers to form a patch. For each patch, random Gaussian noise with zero mean and $0.03$ standard deviation is applied to corrupt the point clouds. Therefore, for each point cloud, we perturb approximately 400 points out of 1024 points.   

\subsection{SeRP-PointNet}

We modify the architecture of PointNet \cite{pointnet} auto-encoder to form SeRP-PointNet. We use PointNet architecture to learn a global 1024-dimensional vector representation of the perturbed point cloud. Similar to PointNet's segmentation network, the global vector representation is concatenated with the local features followed by 4 fully-connected layers to give per-point 128-dimensional features. To pre-train the SeRP-PointNet auto-encoder we use two tasks:
\begin{itemize}
    \item \textbf{Classification:} Classify which points are perturbed. It is performed by a classification head consisting of a fully-connected layer on top of the per-point features. We use a Cross-entropy loss function.
    \item \textbf{Reconstruction:} Reconstructing the perturbed points. We attach a fully-connected layer over the top of per-point features to form the reconstruction head. We provide two methods to learn reconstruction:
    \begin{enumerate}
        \item $\delta$-learning: The reconstruction head predicts $\delta$, i.e. the 3-coordinate differences between the ground truth point cloud and the perturbed point cloud. To calculate the loss, we use a Mean Squared Loss function.
        \item $cd\ell_2$-learning: The reconstruction head predicts the 3-d coordinates directly from the per-point features. For this method, we use the Chamfer Distance $\ell_2$ loss described in Section \ref{subsec:loss_func}.
    \end{enumerate}
\end{itemize}

The total loss is calculated as:

\begin{align}
   \mathcal{L} = \alpha_{cls}*l_{cls} + \alpha_{rec}*l_{rec}
\end{align}

where, $l_{cls}$ and $l_{rec}$ are classification and reconstruction losses respectively. In our experiments, we set $\alpha_{cls}=0.001$ and $\alpha_{rec}=1.5$.

\subsection{SeRP-Transformer}

Transformer models \cite{devlin2018bert} employ self-attention \cite{vaswani2017attention} mechanism to model long-range dependencies between different input tokens. In this section, we will detail our Autoencoder with Transformer backbone and how it is used for processing point cloud data. 

\begin{figure*}
\begin{center}
\includegraphics[width=0.9\linewidth]{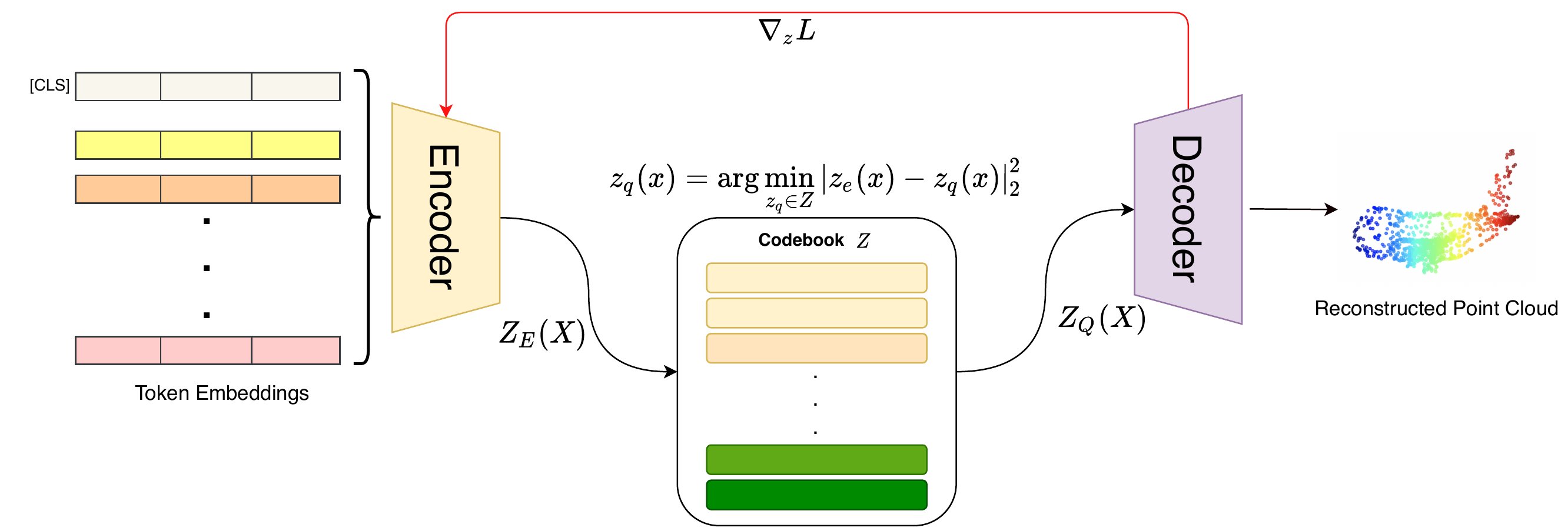}
\end{center}
  \caption{Overview of VASP. Vector Quantization of latent representations is done by maintaining a codebook $\mathcal{Z}$ of learnable discreet latent vectors. The output of the encoder $z_e(x)$ is passed through the codebook that returns the vector $z_q(x)$ closest to $z_e(x)$. As this operation is not differentiable, the gradients of the decoder are copied to the encoder during backpropagation. }
\label{fig:vasp-transformer}
\end{figure*}

\subsubsection{Point Tokenization}
Since the input set of points is large, it becomes computationally infeasible to use transformers based architectures because of the quadratic scaling of self-attention modules. To overcome this, we divide the corrupted point cloud into irregular patches via Farthest Point Sampling (FPS) and K-Nearest Neighbors (KNN) algorithm. Formally, given an input point cloud with $p$ points called $X \in \mathbb{R}^{p\times 3}$, we sample a set of $c$ centers called C, and then KNN is applied on those centers to form patches called the patch tokens, T containing $n$ points in each patch.
\begin{align}
    C &= FPS(X), \ C \in \mathbb{R}^{c\times 3}\\
    T &= KNN(C), \ T \in \mathbb{R}^{c\times n \times 3}
\end{align}
Furthermore, each of the patch token is normalized with respect to the mean and standard deviation of the points present in each patch to improve convergence.

\subsubsection{Embedding} 
To learn the patch token embeddings, $T_e$, we use a lightweight PointNet architecture. Since the patches are center normalized, the position embeddings of the center coordinates are also appended to $T_e$. The position embeddings, $P_e$ are learned using a small MLP networks. Both $T_e$ and $P_e$ are of same dimensions, $d$ and $T_e$ and $P_e$ are concatenated to form the inputs embeddings $T_i$
\begin{align}
    T_e &= \text{PointNet}(T), \ T_e \in \mathbb{R}^{c \times d} \\
    P_e &= \text{MLP}(C), \ P_e \in \mathbb{R}^{c \times d} \\
    T_i &= \text{Concatenate}[T_e; P_e]  \in \mathbb{R}^{c \times 2d}
\end{align}
Following standard transformers \cite{devlin2018bert}, we also append a learnable [CLS] token and its learnable position embedding with the patch tokens $T_i$. The [CLS] token representations learns the overall point cloud structure and it is used in the downstream evaluation tasks. Therefore, the input, $I$, to the auto-encoder is the set of $c+1$ vectors each of dimensions $2d$. 
\begin{align}
    I &= ([CLS]; T_i), \ I \in \mathbb{R}^{(c+1) \times 2d}
\end{align}

\subsubsection{Encoder-Decoder}
Our Autoencoder is based on standard Transformers with asymmetric encoder-decoder design as in \cite{point-mae}. Our transformer encoder takes the inputs and passes it through the self-attention layers to form low-dimensional representations of size $\ell$, $E$. The decoder, then takes those representations and projects them back in the original subspace to give $D$. 
\begin{align}
    E &= \text{Encoder}(I), \ E \in \mathbb{R}^{(c+1) \times \ell} \\
    D &= \text{Decoder}(E), \ D \in \mathbb{R}^{(c+1) \times d}
\end{align}
\subsubsection{Reconstruction Head}
The reconstruction is a fully-connected layer, FC, that outputs the point reconstructions. The decoder output, $D$, is fed to FC to output $\hat{T} \in \mathbb{R}^{c \times 3n}$ where $\hat{T}$ consists of $c$ vectors corresponding to each input patch token, and each reconstructed patch consists of $n$ generated points, each with $3$ coordinates. 
\begin{align}
    \hat{T} = \text{FC}(D), \ \hat{T} \in \mathbb{R}^{c \times 3n}
\end{align}

We use Chamfer Distance $\ell_2$ \cite{chamfer} reconstruction loss as described in Section \ref{subsec:loss_func}. 

The complete overview of the architecture is described in figure \ref{fig:serp-transformer}.

\begin{figure*}
\begin{center}
\includegraphics[width=0.9\linewidth]{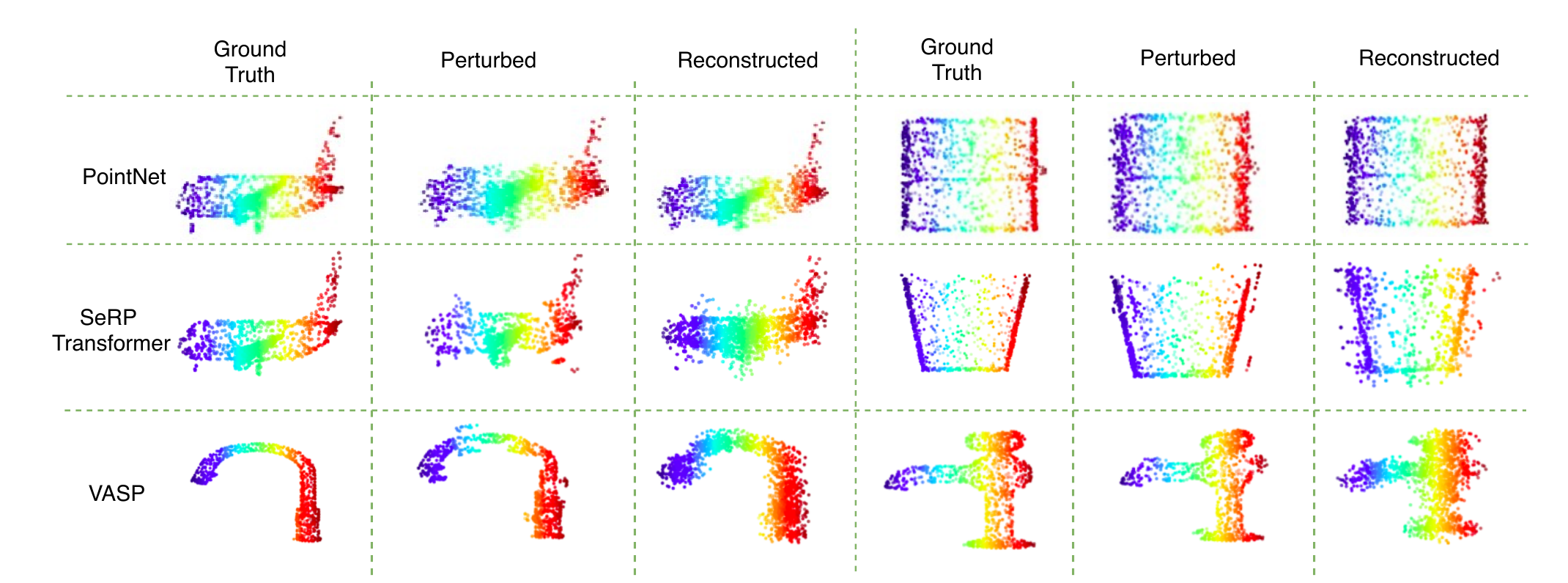}
\end{center}
   \caption{Reconstructions of the ShapeNet validation set: Above figure shows some examples of the original point cloud, corrupted point cloud and the reconstructed point by all the three autoencoders that we proposed. We can see that all the models are able to recover the shape from the latent representations.}
\label{fig:reconstructions}
\end{figure*}

\subsubsection{Loss function}
\label{subsec:loss_func}

\textbf{Chamfer Distance $\ell_2$  Loss:} We use reconstruction loss given by $\ell_2$ Chamfer Distance \cite{chamfer}, where given an input point cloud, $P$ and reconstructed point cloud $\hat{P}$, the loss is given by 
\begin{align}
\label{eq:chamfer}
    \mathcal{L}(P, \hat{P}) = \frac{1}{|\hat{P}|}\sum_{x \in \hat{P}} \min_{y \in P}\| x - y \|_2^2 + \frac{1}{|P|}\sum_{x \in P} \min_{y \in \hat{P}}\| x - y \|_2^2
\end{align}

\subsection{VASP: Vector-Quantized Autoencoder for Self-supervised Representation Learning for point Clouds}
\label{sec:vasp}

Generally, lots of real-world data can be modeled by discrete representations like language, human speech as well as images. Furthermore, architectures like Transformers are designed to work on discrete datasets. Therefore, in order to extend the SeRP-Transformer autoencoder for doing discrete variational inference, we introduce a codebook, $\mathcal{Z}$ that contains a set of $K$ discrete latent vectors, $e_1, \dots, e_K$, of dimension $d$ which is same the dimensions of encoder output. As shown in Figure \ref{fig:vasp-transformer}, the output of the encoder, $z_e(x)$, are passed through this codebook and the discrete latent variable, $z_q(x)$, is computed by the nearest neighbor look-up in the embedding space. 

\begin{align}
    z_q(x) &= e_k \quad \text{where } k = \arg\min_j ||z_e(x) - e_j||^2
\end{align}

The latent vector, is then passed through the decoder for reconstruction. 

\textbf{Learning:} Note that, selecting the latent vector from the codebook, is not a differentiable operation. Therefore, during backpropagation, the gradients of the decoder are copied directly to the encoder. Since the outputs of the encoder and the inputs of the decoder are in the same representational space, the gradients contain usefull information about how the encoder weights should be updated for reconstruction \cite{vq-vae}. The full objective is given by equation \ref{eq:vasp-loss}

\begin{equation}
\label{eq:vasp-loss}
    \mathcal{L}_{VQ} = \log p(x | z_q(x)) + \alpha ||sg[z_e(x)] - e||^2_2 + \beta ||z_e(x) - sg[e]||^2_2
\end{equation}

where $sg[.]$ denotes the stop gradient operation. The first term is the reconstruction loss given by L2 Chamfer distance given equation \ref{eq:chamfer}. The second term of the equation \ref{eq:vasp-loss} makes the embeddings closer to the encoder output and thus, induces learning of the discrete latent space without updating $z_e(x)$. The third term is called the commitment loss function because it makes the output of the encoder commit to an embedding and the embedding is not updated in this term because of the stop-gradient operation. The commitment loss is added because the output space of the encoder is not constrained and thus, the encoder outputs can keep growing if the discrete embeddings are not learned as fast the encoder representations. 

\section{Experiments and Results}
For SeRP-Transformer, we use $c=64$ centers and each center consists of $n=32$ nearest neighbors. The value of $d$ was set to $384$, therefore, the input $T_i$ was $786$ dimensions and latent representations size was set to $\ell=384$. 
\subsection{Pretraining Setup}
We use ShapeNet \cite{shapenet} dataset for pre-training our autoencoders. ShapeNet dataset is a large point cloud dataset containing approximately 50,000 models spanning across 55 categories. The dataset contains dense points clouds and for training we sample 1024 points using Farthest Point Sampling (FPS) as the input point cloud on which we apply perturbation before feeding it to the auto-encoder. Training was done using AdamW optimizer with initial learning rate of $0.001$ with cosine annealing with a batch size of 128. SeRP-PointNet was trained for $100$ epochs, whereas SeRP-Transformer was trained for $300$ epochs. 

The reconstruction results are shown in Figure \ref{fig:reconstructions} for all the three autoencoders - SeRP-PointNet, SeRP-Transformer and VASP Transformer. In the figure, we can see that the autoencoders are able to recover the shape of the original point cloud approximately, and therefore, indicating learning of informative latent representations.  

\subsection{Downstream Evaluation}

We evaluate the quality of the learned latent representations by taking the pre-trained encoder and finetuning it on another dataset for classification. We finetune our pre-trained encoder on ModelNet40 dataset which contains CAD models of 40 categories. 

\subsubsection{Downstream Evaluation Results}
We present the results of our downstream evaluation in Tables \ref{tab:eval_transformers} and \ref{tab:eval_pointnet}. From the results, we observe that the encoders of SeRP-PointNet and SeRP Transformer outperform the non-pre-trained models i.e., the models trained from scratch on ModelNet40. 

From the table, we can also see that the VASP Transformer shows worse results than the non-pre-trained encoder, indicating that the encoder, trained using discrete codes, is not able to learn robust representations and therefore, necessitating more work towards discrete representation learning of point clouds. 

\section{Conclusion and Future Work}

In this work, we presented a self-supervised learning framework to learn representations of point cloud data. We trained autoencoder models that aim to reconstruct the original point cloud using perturbed point clouds as inputs and thus, learning low-dimensional latent space in the process. We showed that pre-trained models perform better than the encoders trained from scratch on a downstream classification task. 

For the future work, we can propose a more challenging strategy to perturb point clouds by sampling centers using FPS instead of random sampling, that would encourage corruption at the edges than inside the volume. We can also compare the existing approach with traditional variational inference and discrete variational inference methods. 


\begin{table}[]
\centering
\begin{tabular}{|p{1.5cm}|cc|cc|}
\hline
\multicolumn{1}{|p{1.5cm}|}{\multirow{2}{*}{\diagbox[width=1.9cm]{\small{\textbf{Model}}}{\small{\textbf{Dataset}}}}} & \multicolumn{2}{c|}{\textbf{ModelNet40}} & \multicolumn{2}{c|}{\textbf{ShapeNet}} \\ \cline{2-5} 
\multicolumn{1}{|l|}{}                     & \multicolumn{1}{c|}{Accuracy}   & Gain   & \multicolumn{1}{c|}{Accuracy}  & Gain  \\ \hline
\textbf{Scratch}                           & \multicolumn{1}{c|}{88.48}      & -      & \multicolumn{1}{c|}{87.43}     & -     \\ \hline
\textbf{SeRP-Net}                          & \multicolumn{1}{c|}{\textbf{89.1}}      & 0.62 \color{ForestGreen} $\uparrow$   & \multicolumn{1}{c|}{\textbf{87.97}}     & 0.54 \color{ForestGreen} $\uparrow$  \\ \hline
\textbf{VASP}                              & \multicolumn{1}{c|}{87.85}      & -0.63 \color{red} $\downarrow$  & \multicolumn{1}{c|}{86.54}     & -0.89 \color{red} $\downarrow$ \\ \hline 
\end{tabular}
\caption{Downstream Task Evaluation for SeRP-Transformer variants. \textit{\small{Scratch = Non-pre-trained model.}} }
\label{tab:eval_transformers}
\end{table}

\begin{table}[]
\centering
\begin{tabular}{|p{1.5cm}|cc|cc|}
\hline
\multicolumn{1}{|p{1.5cm}|}{\multirow{2}{*}{\diagbox[width=1.9cm]{\small{\textbf{Model}}}{\small{\textbf{Dataset}}}}} & \multicolumn{2}{c|}{\textbf{ModelNet40}} & \multicolumn{2}{c|}{\textbf{ShapeNet}} \\ \cline{2-5} 
\multicolumn{1}{|l|}{}                     & \multicolumn{1}{c|}{Accuracy}   & Gain   & \multicolumn{1}{c|}{Accuracy}  & Gain  \\ \hline
\textbf{Scratch}                           & \multicolumn{1}{c|}{82.97}      & -      & \multicolumn{1}{c|}{84.24}     & -     \\ \hline
\textbf{$\delta$ learn}                          & \multicolumn{1}{c|}{\textbf{84.1}}      & 1.13 \color{ForestGreen} $\uparrow$   & \multicolumn{1}{c|}{\textbf{84.43}}     & 0.19 \color{ForestGreen} $\uparrow$  \\ \hline
\textbf{$cd\ell_2$ learn}                              & \multicolumn{1}{c|}{84.06}      & 1.09 \color{ForestGreen} $\uparrow$  & \multicolumn{1}{c|}{84.39}     & 0.15 \color{ForestGreen} $\uparrow$ \\ \hline 
\end{tabular}
\caption{Downstream Task Evaluation for SeRP-PointNet variants. \textit{\small{Scratch = Non-pre-trained model.}}}
\label{tab:eval_pointnet}
\end{table}

{\small
\bibliographystyle{ieee_fullname}
\bibliography{egbib}
}

\end{document}